\documentclass[10pt,twocolumn,letterpaper]{article}

\usepackage[pagenumbers]{cvpr} 

\usepackage{graphicx}
\usepackage{amsmath}
\usepackage{amssymb}
\usepackage{booktabs}
\usepackage{mathtools}
\usepackage{multirow}
\usepackage{array, makecell} %
\usepackage{enumitem}

\usepackage{amsmath,amsfonts,bm}

\def\eqref#1{equation~\ref{#1}}

\def\1{\bm{1}}

\def\vv{{\bm{v}}}

\DeclareMathAlphabet{\mathsfit}{\encodingdefault}{\sfdefault}{m}{sl}
\SetMathAlphabet{\mathsfit}{bold}{\encodingdefault}{\sfdefault}{bx}{n}

\newcommand{\R}{\mathbb{R}}

\newcommand{\thetav}{\boldsymbol{\theta}}
\newcommand{\phiv}{\boldsymbol{\phi}}

\newcommand{\phiw}{\boldsymbol{w}}

\newcommand{\tv}[0]{{\boldsymbol{t}}}

\newcommand{\Xcal}{\mathcal{X}}

\newcommand{\Bcal}{\mathcal{B}}

\newcommand{\Tcal}{\mathcal{T}}

\newcommand{\Lcal}{\mathcal{L}}

\newcommand{\Scal}{\mathcal{S}}

\newcommand{\sv}{{\boldsymbol{s}}}
\newcommand{\uv}{\boldsymbol{u}}
\newcommand{\pv}{\boldsymbol{p}}

\newcommand{\xv}{\boldsymbol{x}}

\usepackage{xcolor}
\usepackage{pifont}
\usepackage[pagebackref,breaklinks,colorlinks]{hyperref}

\usepackage[capitalize]{cleveref}
\crefname{section}{Sec.}{Secs.}
\Crefname{section}{Section}{Sections}
\Crefname{table}{Table}{Tables}
\crefname{table}{Tab.}{Tabs.}
\usepackage{caption}
\usepackage{subcaption}

\def\cvprPaperID{7044} 
\def\confName{CVPR}
\def\confYear{2023}
\def\ours{Pic2Word\xspace}

\begin{document}

\title{RIT: Representing Image as Token for Zero-shot Composed Image Retrieval}

\title{Pic2Word: Mapping Pictures to Words for Zero-shot Composed Image Retrieval}

\author{%
  Kuniaki Saito$^{1,2}$\thanks{Work done during internship at Google Cloud AI Research.}\:,\; Kihyuk Sohn$^{3}$\:,\; Xiang Zhang$^{2}$\:,\; Chun-Liang Li$^{2}$\:,\; \\
  {Chen-Yu Lee$^{2}$\:,\;Kate Saenko$^{1,4}$\:,\;Tomas Pfister$^{2}$}\\
  \small{\texttt{\{keisaito, saenko\}@bu.edu}} \\
  \small{\texttt{\{kihyuks,fancyzhx,chunliang,chenyulee,tpfister\}@google.com}}\\[.2cm]
  $^{1}$Boston University, $^{2}$Google Cloud AI Research, $^{3}$Google Research, $^{4}$MIT-IBM Watson AI Lab\\
}

\maketitle

\begin{abstract}

In Composed Image Retrieval (CIR), a user combines a query image with text to describe their intended target. 
Existing methods rely on supervised learning of CIR models using labeled triplets consisting of the query image, text specification, and the target image. Labeling such triplets is expensive and hinders broad applicability of CIR. 
In this work, we propose to study an important task, Zero-Shot Composed Image Retrieval (ZS-CIR), whose goal is to build a CIR model without requiring labeled triplets for training. 
To this end, we propose a novel method, called \ours, that requires only weakly labeled image-caption pairs and unlabeled image datasets to train. 
Unlike existing supervised CIR models, our model trained on weakly labeled or unlabeled datasets shows strong generalization across diverse ZS-CIR tasks, \eg, attribute editing, object composition, and domain conversion. Our approach outperforms several supervised CIR methods on the common CIR benchmark, CIRR and Fashion-IQ. Code will be made
publicly available at \url{https://github.com/google-research/composed_image_retrieval}

\end{abstract}

\section{Introduction}
\label{sec:intro}

\begin{figure}
    \centering
    \includegraphics[width=0.99\linewidth]{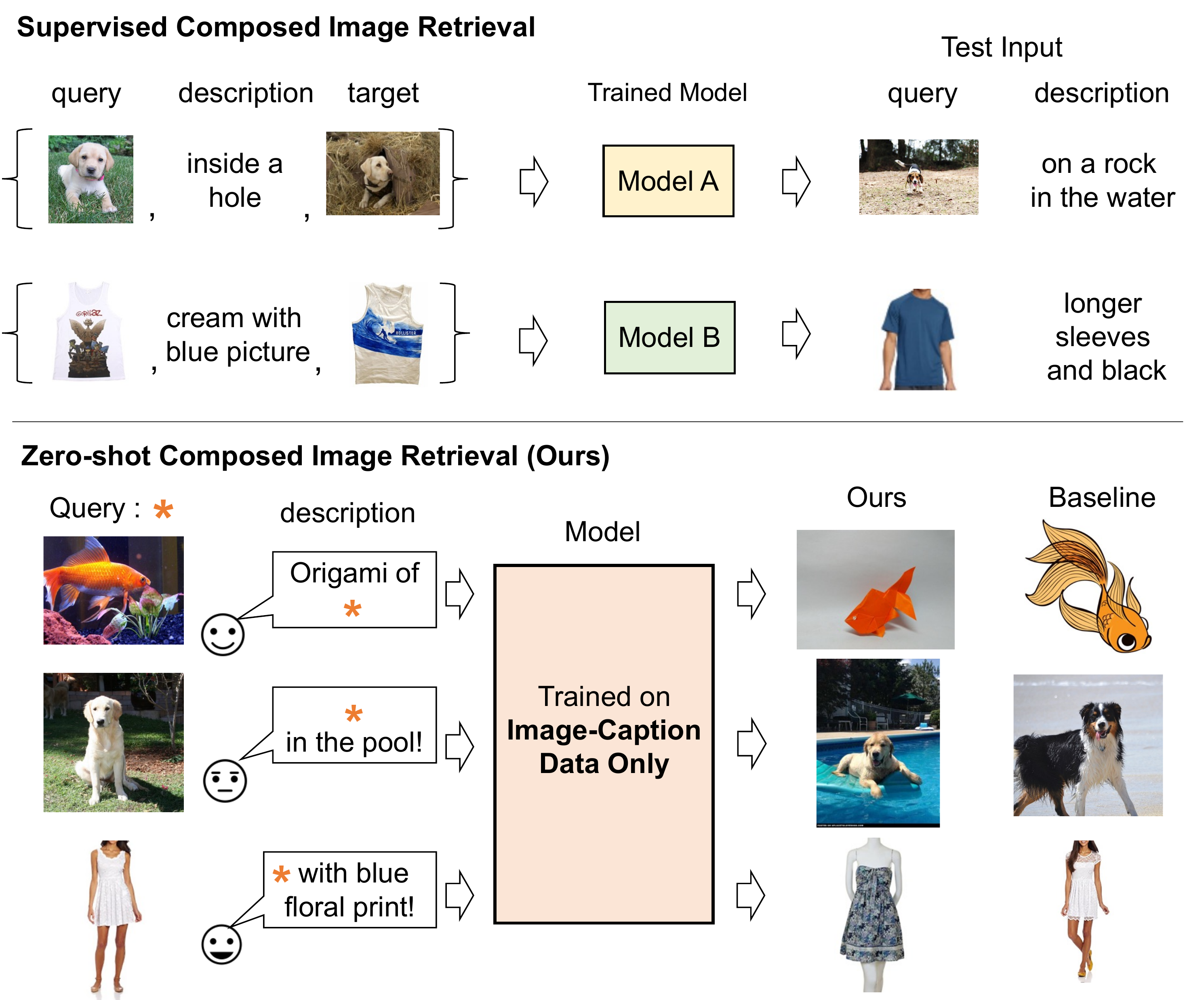}
     \vspace{-2mm}
    \caption{Composed Image Retrieval (CIR) takes a query composed of an image (denoted as {\color{orange} $\star$}) plus a text modifier and retrieves matching images. CIR encompasses diverse tasks, such as domain conversion (origami of {\color{orange} $\star$}), scene or object composition ({\color{orange} $\star$} in the pool), or fashion-attribute manipulation ({\color{orange} $\star$} with blue floral print). \textbf{Top}: Existing methods~\cite{vo2019composing,baldrati2022effective} train a separate model for each task, and require strong triplet supervision. \textbf{Bottom}: We propose a new task, Zero-shot CIR, and solve diverse CIR sub-tasks  using a single model trained only on image-caption pairs and unlabeled image datasets.}
    \label{fig:fig1}
\end{figure}

\begin{figure*}
    \centering
    \includegraphics[width=\textwidth]{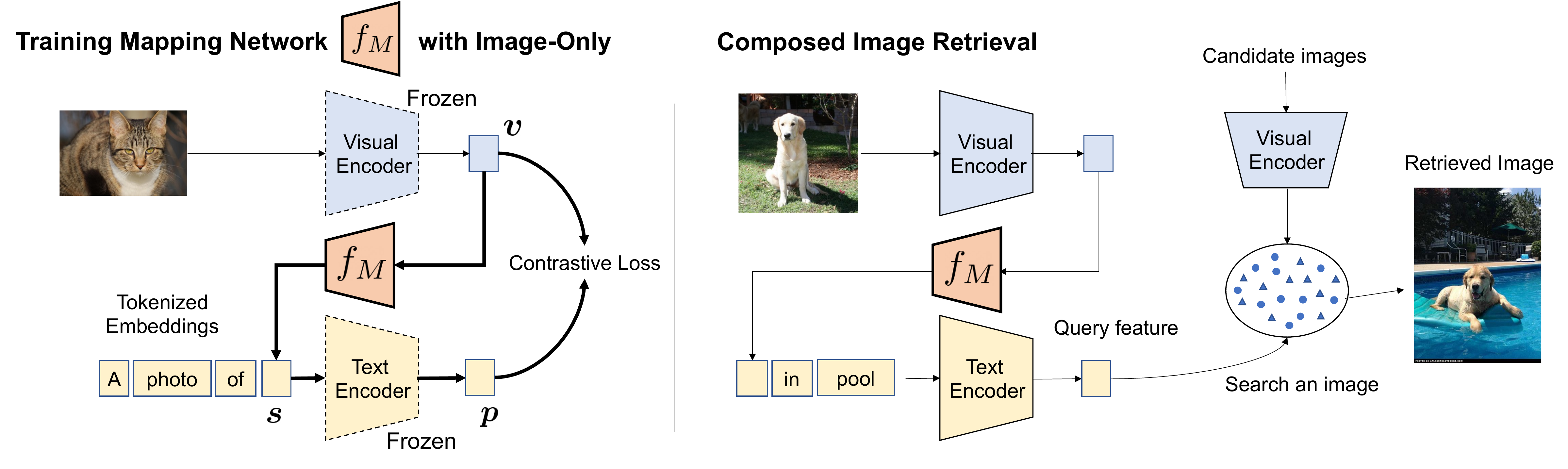}
    \vspace{-7mm}
    \caption{\textbf{Left}: Overview of training our \ours mapping network. Given a frozen visual and text encoder, the mapping network, $f_{M}$, is optimized to minimize the contrastive loss between the image embedding and the language embedding generated by the pseudo word token $s$. \textbf{Right}: Overview of inference. 
    The estimated token is used to fill in the given prompt sentence.}
    \label{fig:pipeline}
\end{figure*}

{
Composed image retrieval (CIR)~\cite{vo2019composing,chen2020image,hosseinzadeh2020composed,lee2021cosmo,baldrati2022effective} aims to retrieve images using a query composed of an image and text. In contrast to the image-based retrieval systems~\cite{datta2008image}, CIR is advantageous as it allows retrieval of images with a higher precision thanks to the text query that incorporates user's intent, such as a desired modification to the query image. With a surge of image-text models~\cite{radford2021learning,li2021align,alayrac2022flamingo}, CIR has received  attention recently for diverse real-world applications in e-commerce and internet search.

Several approaches~\cite{baldrati2022effective,vo2019composing,lee2021cosmo,chen2020image,hosseinzadeh2020composed} have been proposed to solve CIR problems, including attribute manipulation for fashion image search, composing objects, and converting the style of images for content creation, as in \cref{fig:fig1}. At the core of CIR is learning how to compose information from an image and text. We identify two fundamental issues with existing solutions. 
First, previous methods require a large amount of labeled data, which comes in the form of triplets consisting of a reference image, text, and a target image, to train their retrieval model. The dataset collection involves two processes~\cite{liu2021image} – collecting pairs of a related reference and the target images as a query-output pair to the CIR system, then giving a description that modifies the reference to the target. Examples of labeled triplets are at the top of \cref{fig:fig1}. We note that both steps incur a significant labeling cost.
Second, the model trained on labeled data is specialized to specific use-cases and may not generalize to different CIR tasks.

To tackle these challenges, we propose a new task, \emph{zero-shot composed image retrieval} (ZS-CIR). In ZS-CIR, our goal is to build a single CIR model that performs diverse CIR tasks, such as object composition, attribute editing, or domain conversion, as in the bottom of \cref{fig:fig1}, without requiring an expensive labeled triplet data collection effort. Instead, we propose to train our retrieval model using large-scale image-caption pairs and unlabeled images, which are considerably cheaper to collect than supervised CIR datasets at scale. 

To harness these weakly labeled and unlabeled datasets, we propose a two-stage framework for learning ZS-CIR models. The first stage  conducts contrastive language-image pretraining (CLIP)~\cite{radford2021learning} on the image-caption dataset, training a two-tower model jointly to maximize the similarity between an image and a caption. We note that some previous works~\cite{baldrati2022effective} build their models on CLIP followed by a second-stage supervised CIR training.
%
%
On the contrary, instead of relying on the triplet-labeled training data, we leverage the linguistic capability of the language encoder in CLIP, which excels at composing diverse concepts or attributes to generate embeddings that are close to the corresponding visual representations. The idea is to map a \emph{picture to a word token} such that the language encoder can flexibly and seamlessly compose the query image features and text descriptions. We learn a lightweight mapping network that converts an image embedding of the CLIP vision encoder into a token embedding of its language encoder. The mapping network is trained with a contrastive loss to reconstruct the image embedding, which only requires unlabeled images.  We call our method \ours and illustrate it in \cref{fig:pipeline}.

In experiments, we show the strength of our \ours approach on various CIR tasks, including domain conversion, object composition, scene manipulation, and fashion attribute manipulation. 
We show that our zero-shot approach performs on-par with or better than several recent supervised CIR methods~\cite{delmas2022artemis,liu2021image,dodds2020modality} relying on labeled training data.
Our contributions are threefold:
\vspace{-3mm}
\begin{itemize}[noitemsep]
    \item We propose a new task, zero-shot composed image retrieval (ZS-CIR), which aims to solve diverse CIR tasks without requiring expensive triplet-labeled training datasets. 
    \item We propose \ours, a novel method for ZS-CIR that is trainable using only image-caption and unlabeled image datasets. 
    \ours leverages pre-trained vision-language models and transforms an input image to a language token in order to flexibly compose image and text queries. 
    \item \ours improves the ZS-CIR performance, e.g., relative improvement of 10 - 100\% on four CIR tasks, which is on-par with several recent CIR methods using labeled training data.
\end{itemize}
}
\section{Related Work}
\textbf{Composed Image Retrieval.}
Composed image retrieval (CIR) is proposed to retrieve images with a pair consisting of a reference image and text~\cite{vo2019composing}. The topic has been explored in the field of fashion~\cite{wu2021fashion} and scene composition~\cite{liu2021image}.
Current state-of-the-art CIR models rely on late-fusion, i.e., combining visual and language features after extracting them with different encoders~\cite{baldrati2022effective, chen2020image, chen2020learning, lee2021cosmo, liu2021image,delmas2022artemis}, where a pre-trained CLIP model shows strong performance\cite{baldrati2022effective}. 
Wu~\etal\cite{wu2021fashion} train a shallow transformer from scratch to fuse image and language features at the input-level. 
Goenka~\etal\cite{goenka2022fashionvlp} utilize a pre-trained BERT~\cite{devlin2018bert} model to fuse image, text, and tag features while Han~\etal\cite{han2022fashionvil} leverage a large-scale fashion dataset to pre-train vision-language model by early fusion, but both still need to train on a CIR dataset. 
These approaches employ the compositional ability of the pre-trained language model by tuning their transformer model on a CIR dataset. Unlike these methods, our approach does not need a CIR dataset for training, yet can handle diverse scenarios.

\textbf{Vision-language foundation models.}
Vision language models such as CLIP~\cite{radford2021learning}, ALIGN~\cite{li2021align}, \etc, pre-train an image and language encoder pair on large-scale data containing hundreds of millions of image-caption pair.
Following CLIP, many vision-language “foundation" models incorporate more data into training, introduce new architectural designs, or utilize new objectives~\cite{li2021align, singh2021flava, yu2022coca,alayrac2022flamingo,jia2021scaling}. 
Since these models have seen varying text describing the concepts in an image, they can handle diverse tasks, e.g., caption-based image retrieval, zero-shot classification~\cite{radford2021learning}, few-shot classification~\cite{zhou2022conditional}, image-captioning~\cite{mokady2021clipcap}, and VQA~\cite{song2022clip}, with a small or no extra annotation cost. In this work, we present an approach to adapt the foundation models to CIR tasks without the need for a CIR dataset. We are the first to apply vision-language contrastive learning models to CIR in a zero-shot manner.

\textbf{Representing image as one word.}    
Several approaches attempt to represent an image region as a token during the pre-training of vision-language models~\cite{wang2022vlmixer, chen2019uniter, li2020oscar, lu2019vilbert}. The typical framework consists of (i) detecting objects in an image using a pre-trained object detector, (ii) feeding the regions and sentence to a text encoder, and (iii) optimizing several multi-modal objectives to obtain a strong vision-language model. These approaches require a high-performing object detector in the  pre-training stage, while we present an approach applicable to pre-trained contrastive vision-language models. 
A concurrent work~\cite{cohen2022my} learns a network with cycle contrastive loss to tackle different problems, personalized image retrieval and semantic segmentation. To achieve those, they learn to transform a \emph{set of images} into a concept word token, which requires images with class-wise annotations and caption annotations. In addition, they require fine-tuning the word token given a new concept consisting of a few images. Instead, we focus on composed image retrieval. The proposed \ours does not need such annotations or fine-tuning the word token on few-shot labeled images.  
Gal \etal~\cite{gal2022image} propose to represent a few images depicting the same object with a word token and use this for a text-to-image generation task. Our approach does not require such a group of images, nor any training at inference time. We believe our approach can potentially be applied to image generation in future work.

\section{Method}
In this section, we introduce our approach for ZS-CIR, \ours. The overview of the training and the inference is described in Fig.~\ref{fig:pipeline}.
We utilize a  pre-trained image-text model, CLIP~\cite{radford2021learning}, that consists of a language and a vision encoder. The output embeddings of two encoders are aligned with respect to each other's modality. 
Given a frozen pre-trained model, we train a mapping network to convert the visual embedding into the corresponding \emph{pseudo language token}. 
The network is optimized such that the pseudo token represents the visual embedding faithfully. At test time the predicted token is simply inserted into a template together with the query text and the resulting features are compared to candidate images. We detail the training and inference mechanisms in the following subsections. 

\subsection{Preliminaries}
\label{sec:preliminary}
\vspace{-1mm}
\noindent\textbf{Contrastive Language-Image Pre-training (CLIP).}
Suppose we have access to an image-caption dataset with image-caption pairs  $\Scal  \,{=}\,\{ (\xv_n, \tv_n) \}_{n=1}^{N}$, where $\xv \,{\in}\, \Xcal$  is an image and $\tv \,{\in}\, \Tcal$ is a tokenized language description.
For an image $\xv$, an image encoder model $f_{\thetav}$ parameterized by $\thetav$ extracts a visual representation $\Tilde{\vv}  \,{\in}\, \R^{d\times 1}$: $ \Tilde{\vv} \,{=}\, f_{\thetav}(\xv)$. For each text description $\tv \,{\in}\, \Tcal$ (e.g., caption or prompt), a word embedding layer $E_{\phiw}$ parameterized by $\phiw$ extracts the \emph{token embedding} $E_{\phiw}(\tv)$. We will use the terms \emph{token embedding} and \emph{token} interchangeably for concise presentation when there is no ambiguity. Then a text encoder $f_{\phiv}$ parameterized by $\phiv$ extracts a language representation $ \Tilde{\uv}  \,{\in}\, \R^{d \times 1}$ following $E_{\phiw}(\tv)$ as $ \Tilde{\uv}  \,{=}\, f_{\phiv}(E_{\phiw}(\tv))$. 

CLIP~\cite{radford2021learning} is designed to find representations that match an image to its paired caption while separating unpaired ones.
For $i$-th image $\xv_i$ and $j$-th language description $\tv_j$ in a batch $\Bcal$, their features are normalized using $ \vv_i\,{ =}\, \frac{  \Tilde{\vv}_i   }{ \| \Tilde{\vv}_i \| }  $ and $ \uv_j \,{=}\, \frac{    \Tilde{\uv}_j }{   \| \Tilde{\uv}_j  \|} $. Finally, CLIP optimizes the symmetric multi-class N-pair loss \cite{sohn2016improved}:
\vspace{-2mm}
\begin{align}\label{eq:obj_bicon}
	\min_{ \{ \thetav, \phiv, \phiw\} } ~~ \Lcal_{con} 	= & \Lcal_{t2i} + \Lcal_{i2t}, 
\end{align}
which includes two contrastive terms, with a temperature hyper-parameter $\tau$ that controls the strength of penalties on hard negative samples, as follows:
\vspace{-2mm}
\begin{align}
\Lcal_{t2i}	= - \frac{1}{ |\Bcal| }\sum_{ i \in \Bcal }  
\log \frac{ \exp(\tau \uv_{i}^T \vv_i)  }{\sum_{ j \in \Bcal}  \exp(\tau \uv_{i}^T \vv_{j})  }, \label{eq:obj_i2t}\\
\quad \Lcal_{i2t}	= - \frac{1}{ |\Bcal| }\sum_{ i \in \Bcal } 
\log \frac{ \exp(\tau \vv_{i}^T \uv_i )  }{\sum_{j \in \Bcal}  \exp(\tau \vv_{i}^T \uv_{j} )  }.\label{eq:obj_t2i}
\end{align}
Note that image-caption retrieval is performed on the normalized feature space, i.e., $ \vv_i $ and $ \uv_i$. 

\iffalse
To retrieve images using image as a query, we rank the similarity between embeddings of query image and the candidate images and select the ones with the highest similarity. Similarly, one can use CLIP to retrieve images using text as a query, by using a similarity computed between the query text embedding and the candidate image embeddings instead. One may achieve the composed image retrieval using both image and text as query by combining these two types of similarity.

\subsection{Contrastive Learning}

As discussed, CLIP embeddings could be used directly for CIR by composing similarity scores. One caveat is that, since they are composed at a high-level, balancing the similarity scores of image-to-image and text-to-image requires careful tuning and text embedding could easily override the influence from the visual embedding, as shown in \cref{fig:weight_study}.

We seek for a method that can \emph{deeply compose} the query image and text signals for CIR. To this end, we propose to represent an image as a pseudo language token, and leverage the linguistic capability of CLIP language encoder to compose an image, which is now represented as one of language tokens, and text via multiple layers of transformer.

To represent an image as a pseudo language token, we design a mapping network, in \cref{fig:pipeline}. Specifically, given a pretrained vision and language encoder of CLIP, the mapping network $f_{M}$, parameterized by $M$, maps an input visual embedding $\Tilde{\vv}$ into a token $\sv\,{=}\,f_{M}(\Tilde{\vv})$ that can be fed into the language encoder of CLIP. We build a network $f_{M}$ with three fully-connected layers, composed of roughly 0.8M parameters. As in the left of Fig.~\ref{fig:pipeline}, we append $\sv$ at the end of token embeddings of the text prompt, \texttt{a photo of}, resulting in $\hat{\sv}$. Then, we feed $\hat{\sv}$ into the language encoder $f_{\phi}$ to obtain the language embedding $\Tilde{\pv} \,{=}\, f_{\phiv}(\hat{\sv})$. To make sure that $\Tilde{\pv}$ is able to represent an input image embedding $\Tilde{\vv}$, we propose to train the mapping network by minimizing the contrastive loss as follows:
\else

\subsection{Learning the \ours Mapping Network}

We train \ours with contrastive loss to map an input visual embedding $\Tilde{\vv}$ to a pseudo language token that  is compatible with the embedding generated by the frozen CLIP language encoder. In CLIP, the language and vision encoders are optimized to align the language embedding $\uv$ and the image embedding~$\vv$. We hypothesize that a pseudo token represents the image semantics faithfully if a language embedding from a generic prompt plus the pseudo token embedding is close to the corresponding image embedding. 
Following this intuition, we propose to train a mapping network to output pseudo language tokens such that the language embedding obtained by a fixed language encoder can minimize the contrastive loss between the visual and language embeddings, i.e., enforcing the network to form a cycle starting from the visual embedding to the final language embedding as shown in the left of Fig.~\ref{fig:pipeline}.

For an unnormalized visual embedding $\Tilde{\vv}$, the mapping network $f_{M}$ with parameters $M$ extracts a pseudo language token embedding $\sv\,{=}\,f_{M}(\Tilde{\vv})$. We build with a three-layered fully-connected network for $f_{M}$ with roughly 0.8M parameters. 
As shown in the left of Fig.~\ref{fig:pipeline}, we append $\sv$ at the end of token embeddings of the prompt sentence, \texttt{a photo of}, resulting in $\hat{\sv}$. Then, we feed $\hat{\sv}$ into the language encoder $f_{\phiv}$ to obtain the language embedding $\Tilde{\pv} \,{=}\, f_{\phiv}(\hat{\sv})$, hoping that $\Tilde{\pv}$ is able to represent an input image embedding $\Tilde{\vv}$. To achieve this, we propose to minimize the contrastive loss with respect to the mapping network, i.e.,
\fi
\begin{align}\label{eq:obj_bicon_mapping}
	\min_{  M } ~~ \Lcal 	= & \Lcal_{t2i}(\pv, \vv) + \Lcal_{i2t}(\pv, \vv), 
\end{align}
which includes two contrastive terms:
\vspace{-2mm}
\begin{align}
\Lcal_{t2i}(\pv, \vv)	= - \frac{1}{ |\Bcal| }\sum_{ i \in \Bcal }  
\log \frac{ \exp(\tau \pv_{i}^T \vv_i)  }{\sum_{ j \in \Bcal}  \exp(\tau \pv_{i}^T \vv_{j})  },  \\
\quad \Lcal_{i2t}(\pv, \vv)	= - \frac{1}{ |\Bcal| }\sum_{ i \in \Bcal } 
\log \frac{ \exp(\tau \vv_{i}^T \pv_i )  }{\sum_{j \in \Bcal}  \exp(\tau \vv_{i}^T \pv_{j} )  }, 
\end{align}
where $ \pv_i \,{=}\, \frac{    \Tilde{\pv}_i }{   \| \Tilde{\pv}_i  \|} $. In this way, we train $f_{M}$ to generate $\sv_i$ that makes $\pv_i$ close to $\vv_i$. 
Note that these are different from Eq. (\ref{eq:obj_bicon}, \ref{eq:obj_i2t}, \ref{eq:obj_t2i}) in that (i) optimization is done for $M$ while fixing the rest, and (ii) we use $\pv$ in replace of $\uv$. That being said, the mapping network is trained in an unsupervised way using unlabeled images only \emph{without} image-text paired or labeled image data.

\begin{table*}[t]
  \centering \scalebox{0.9}{
  \begin{tabular}{lcrr|rr|rr|rr|rr} 
  \toprule
  \multicolumn{2}{c}{}     &   \multicolumn{2}{c}{Cartoon} & \multicolumn{2}{c}{Origami}& \multicolumn{2}{c}{Toy} & \multicolumn{2}{c}{Sculpture}& \multicolumn{2}{c}{Average}\\
  \cmidrule(lr){3-4}
  \cmidrule(lr){5-6}
  \cmidrule(lr){7-8}
  \cmidrule(lr){9-10}
  \cmidrule(lr){11-12}
  \multicolumn{1}{l}{Supervision} & \multicolumn{1}{l}{Methods} & R10 & R50& R10 & R50& R10 & R50& R10 & R50& R10 & R50\\  \midrule
  \multirow{4}{*}{{\textsc{Zero-shot}}}
  & Image-only  &0.3&4.5&0.2&1.8&0.6&5.7&0.3&	4.0            &0.4&4.0\\
  &  Text-only  &0.2&1.1&0.8&3.7&0.8&2.4&0.4&	2.0&0.5&2.3      \\
  &  Image$+$Text&2.2&13.3&2.0&10.3&1.2&9.7&1.6&	11.6&1.7&11.2                    \\

&  \ours   &\textbf{8.0}&\textbf{21.9}&\textbf{13.5}&\textbf{25.6}&\textbf{8.7}&\textbf{21.6}&\textbf{10.0}&\textbf{23.8}&\textbf{10.1}&\textbf{23.2}              \\
  \cmidrule{1-12}
\multicolumn{1}{c}{CIRR}  &Combiner~\cite{baldrati2022effective} &6.1 &14.8&10.5&21.3&7.0&17.7&8.5&20.4&8.0&18.5\\
\multicolumn{1}{c}{Fashion-IQ}&Combiner~\cite{baldrati2022effective} &	6.0&16.9&7.6&20.2&2.7&10.9&8.0&21.6&6.0&17.4\\
\bottomrule
  \end{tabular}}
  \vspace{-3mm}
  \caption{\textbf{Results of the domain conversion experiment using ImageNet.} \ours outperforms all baselines by a large margin. In particular, \ours outperforms supervised models trained on other composed image retrieval tasks (bottom two rows).}
  \label{tab:imgnet}
  \vspace{-3mm}
\end{table*}

\begin{figure*}
    \centering
    \includegraphics[width=0.95\textwidth]{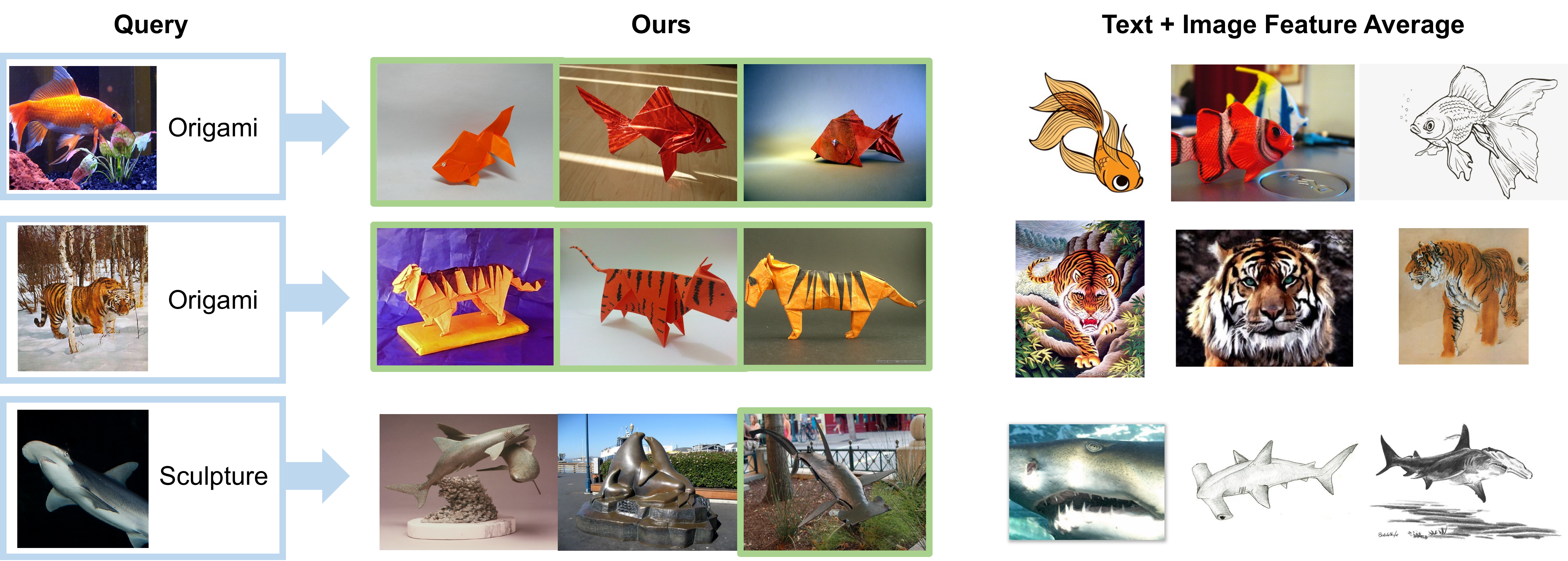}
    \vspace{-3mm}
    \caption{\textbf{Top-3 retrievals in the domain-conversion experiment}. Note that in our approach, the query is composed with the sentence, “a \textit{domain} of *", where \textit{domain} and * is replaced with domain name and image features respectively. The successful results are highlighted with the green outline. Our approach excels at converting the domain of input image features.}
    \label{fig:ex_imgnet}
     \vspace{-3mm}
\end{figure*}

In summary, the mapping network is trained to reconstruct the visual representations in the space of the language embedding. We represent each image as one word token, yet using multiple tokens might be optimal considering the richness of image information. As discussed in Sec~\ref{sec:analysis}, even one word token can represent an image very effectively. We hypothesize that more word tokens may be required to describe very fine details of the image and we can simply extend our idea to obtain multiple word tokens. 
Also, unlike state-of-the-art supervised CIR methods relying on late-fusion~\cite{baldrati2022effective,chen2020image, chen2020learning, lee2021cosmo,liu2021image,delmas2022artemis}, i.e., training a network to combine visual and language features, \ours achieves early-fusion in the language token space. This allows us to harness the capacity of pre-trained language model to compose diverse concepts. Additionally, \ours has the flexibility to accept multiple images to create one query feature since each image is represented as a word (See Fig.~\ref{fig:interesting}).

\subsection{Inference}
\vspace{-2mm}
\label{sec:compose}
At inference our goal is to compose the image and text query and compare it to candidate images.
We compose the pseudo token embedding of the image, from the mapping network, with the text description as follows. 
As shown on the right of Fig.~\ref{fig:pipeline}, the idea is to add a pseudo token into pre-defined prompts as if it were a real word token. The result is embedded by the text encoder and compared to visual features of candidate images.
We note that the prompt design plays an important role of the final performance~\cite{lester2021power,liu2021pre}.  As our focus is on studying the composition of image embeddings and language descriptions in a zero-shot way, we rely on simple prompts without further tuning.
We provide examples of adopted prompts for the different tasks we study.
In all examples, \texttt{[*]} indicates the pseudo token from mapping network.

\textbf{(a) Domain conversion.}
Under the scenario where we want to modify the domain of the query image, e.g., a real image to a sketch-style image, we compose the domain and image as \texttt{a [domain] of [*]}, where \texttt{[domain]} will be replaced by a word specifying the domain.

\textbf{(b) Object/Scene composition.}
In object composition, we aim to retrieve an image consisting of an object specified with a query image, and scene/objects described with text. We compose the query by \texttt{a photo of [*], [obj$_{1}$] and [obj$_{2}$], ..., and [obj$_{n}$]}, where \texttt{[obj$_i$]} are text descriptions of objects or scenes. 

\textbf{(c) Sentence specification.}
The modification to the reference image can be given by a sentence. In such cases, we simply append the sentence after the prompt with a pseudo token such as \texttt{a photo of [*], [text]}, where \texttt{[text]} denotes modification text.

\begin{table}[t]
  \centering 
  \scalebox{0.9}{
  \begin{tabular}{c|c|ccc} 
  \toprule
  \cmidrule{1-5}
  \multicolumn{1}{c}{Supervision} & \multicolumn{1}{c}{Methods} & R1& R5&R10     \\ 
  \midrule
   \multirow{4}{*}{{\textsc{Zero-shot}}} & Image-only    &8.6&15.4&18.9\\ 
   & Text-only   &6.1&15.7&23.5\\
   & Image$+$Text  &10.2&20.2&26.6\\
   & \ours &\textbf{11.5}&\textbf{24.8}&\textbf{33.4}\\
     \cmidrule{1-5}
  \multicolumn{1}{c|}{CIRR} &Combiner~\cite{baldrati2022effective}&9.9&22.8&32.2\\
  \multicolumn{1}{c|}{Fashion-IQ}&Combiner~\cite{baldrati2022effective}&13.2&27.1&35.2\\
  \bottomrule
  \end{tabular}}
   \vspace{-3mm}
  \caption{\textbf{Evaluation on COCO}, i.e., object composition task.}
  \label{tab:coco}
\end{table}

\section{Experiments}
\vspace{-2mm}
In this section, we describe experiments evaluating \ours on zero-shot composed image retrieval. We leverage four datasets to assess the performance of the model in diverse scenarios, including the standard CIR benchmarks CIRR~\cite{liu2021image} and Fashion-IQ~\cite{wu2021fashion}.
This section is organized as follows: (i) an explanation of the experimental setup, (ii) an introduction of the main results, and (iii) a detailed analysis. 

\begin{figure}[t]
    \centering
     \vspace{-3mm}
    \includegraphics[width=0.48\textwidth]{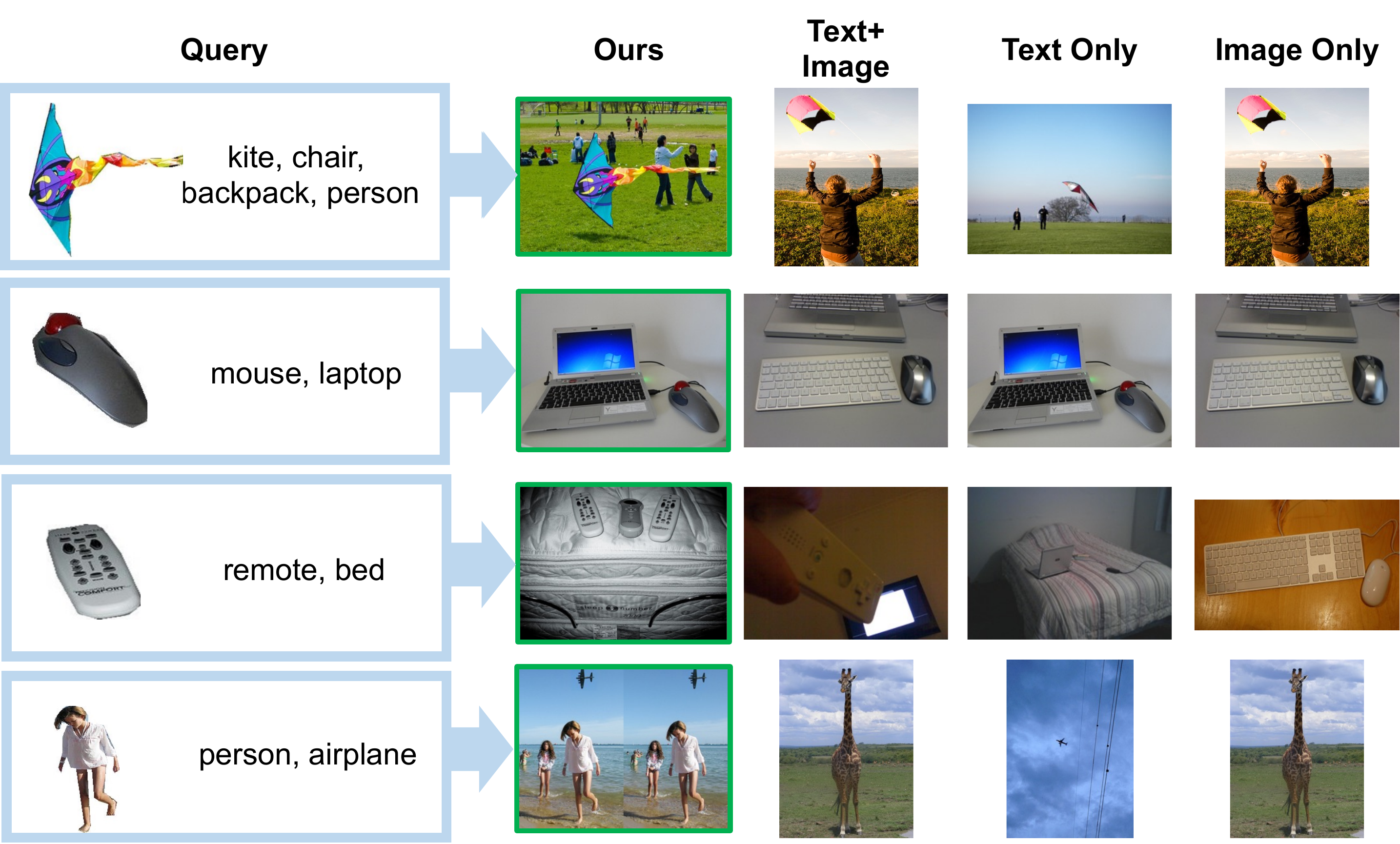}
      \vspace{-6mm}
    \caption{\textbf{Qualitative results on COCO.} We visualize top-1 retrievals of zero-shot methods. Correct retrievals are highlighted with green outline. }
    \label{fig:ex_coco}
\end{figure}
\subsection{Setup}
\textbf{Training Details.}
Unless otherwise stated, we use ViT-L/14 CLIP~\cite{radford2021learning} pre-trained with 400M image-text paired data.\footnote{We employ \href{https://openaipublic.azureedge.net/clip/models/b8cca3fd41ae0c99ba7e8951adf17d267cdb84cd88be6f7c2e0eca1737a03836/ViT-L-14.pt.}{ViT-L/14} published by OpenAI.}
We train the mapping network on the Conceptual Caption dataset~\cite{sharma2018conceptual}, consisting of 3M images. 
The mapping network consists of three-layered MLP of 512 hidden units with ReLU activation (no activation on the output units). We use AdamW~\cite{loshchilov2018decoupled} with $10^{-4}$ learning rate and $0.1$ weight decay. The batch size of the contrastive learning is 1024. The mapping network is trained on 8 Tesla V100 GPUs. We report the performance averaged over three trials.

\noindent\textbf{(a). Domain conversion.}
In this setup, we evaluate the ability to compose domain information and query image representation. 
We utilize ImageNet~\cite{deng2009imagenet} and ImageNet-R~\cite{hendrycks2021many}, which is comprised of 200 classes with diverse domains and has domain annotations. Considering the noise in the annotation, we pick \textit{cartoon, origami, toy and sculpture} as the evaluation target. Then, given the real image from ImageNet and target domain name, we compose the query following the procedure in (a) in Sec~\ref{sec:compose}. 
We regard the retrieved image as correct if its class and domain match with the query image and domain. Due to the lack of datasets that can evaluate instance-level cross-domain retrieval on diverse samples, we rely on  class- and domain-level evaluation. While not ideal this evaluation can still give us good insights into the model's ability to accurately translate the domain of the query image compared to baselines.  
The search is performed on 16,983 images from ImageNet-R and ImageNet, i.e., \textit{cartoon, origami, toy, sculpture and real} images are candidates. 

\begin{table}[t]
  \centering 
  \scalebox{0.9}{
  \begin{tabular}{c|c|cccc} 
  \toprule
  \cmidrule{1-5}
  \multicolumn{1}{l}{Supervision} & \multicolumn{1}{c}{Methods} & R1& R5&R10&R50\\ 
  \midrule
   \multirow{4}{*}{{\textsc{Zero-shot}}} & Image-only    &7.4	&23.6&34.0&57.4\\
   & Text-only   &20.9&44.8&56.7&79.1\\
   & Image$+$Text  &12.4&36.2&49.1&78.2\\
   & \ours &\textbf{23.9}&\textbf{51.7}&\textbf{65.3}&\textbf{87.8}\\
     \cmidrule{1-6}
  \multicolumn{1}{c|}{CIRR} &Combiner~\cite{baldrati2022effective}&30.3&60.4&73.2&92.6\\
  \multicolumn{1}{c|}{Fashion-IQ}&Combiner~\cite{baldrati2022effective}&20.1&47.7&61.6&85.9\\\midrule
  
\multicolumn{1}{c|}{CIRR}&Combiner$^{*}$~\cite{baldrati2022effective}&33.6&65.4&77.4&95.2\\
\multicolumn{1}{c|}{CIRR}&TIRG~\cite{vo2019composing}&14.6&48.4&64.1&90.0\\
   \multicolumn{1}{c|}{CIRR}&ARTEMIS~\cite{delmas2022artemis}&17.0&46.1& 61.3&87.7\\
    \multicolumn{1}{c|}{CIRR}&CIRPLANT~\cite{liu2021image}&19.6&52.6&68.4&92.4\\
  \bottomrule
  \end{tabular}}
   \vspace{-3mm}
  \caption{\textbf{Evaluation on CIRR test set.} Combiner$^{*}$ is the result reported by the authors using ResNet50x4 as a backbone.}
  \label{tab:cirr_test}
\end{table}

\begin{figure}[t]
    \centering
     \vspace{-3mm}
    \includegraphics[width=0.48\textwidth]{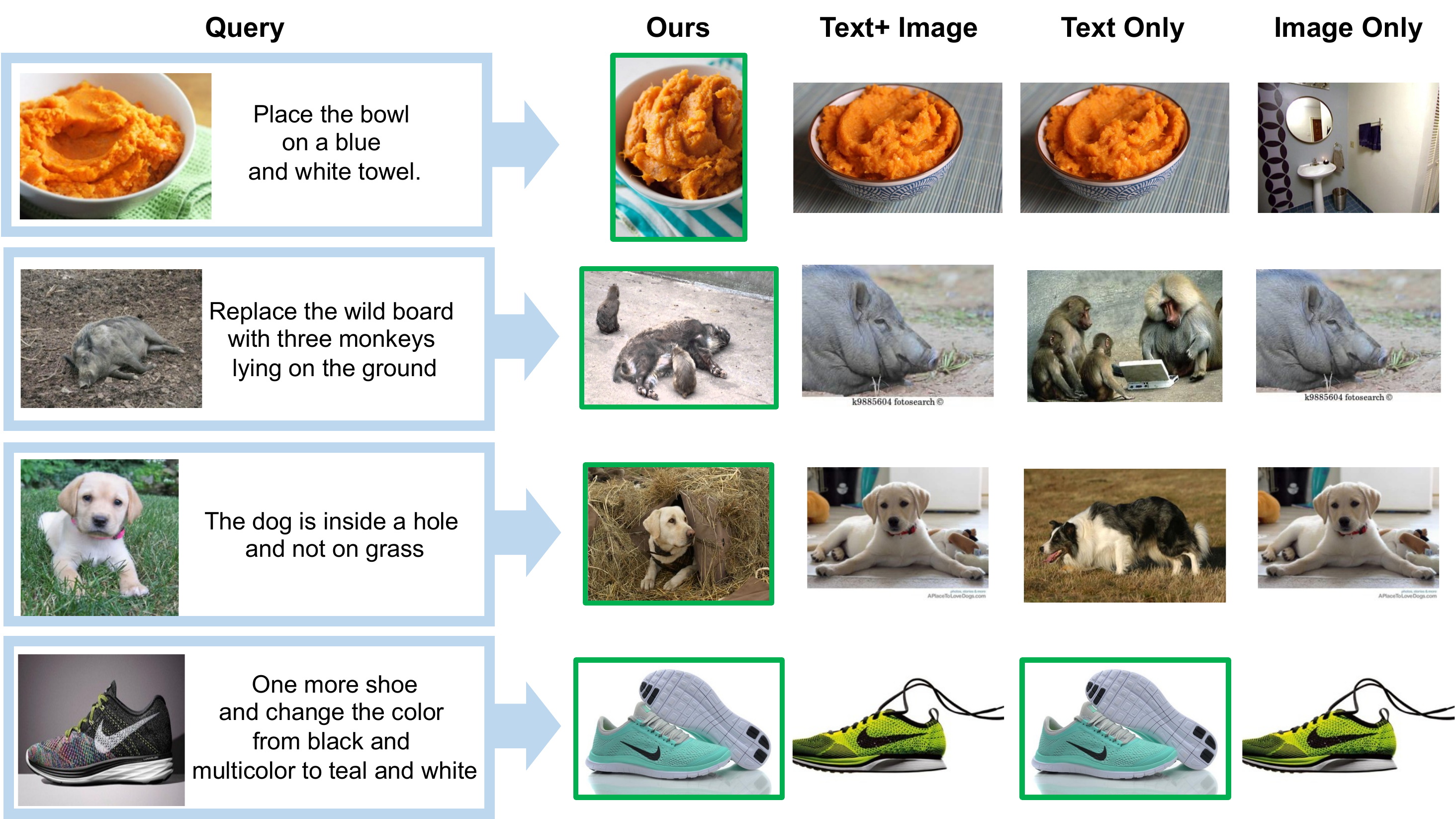}
    \vspace{-5mm}
    \caption{\textbf{Qualitative results on CIRR.} We visualize top-1 retrievals of zero-shot methods. Correct retrievals are highlighted with green outline. We can observe that our approach captures both the object in image and the caption well.} 
    \label{fig:ex_cirr}
\end{figure}


\noindent\textbf{(b). Object composition.}
We evaluate the ability to compose an instance, given as an image, and other scenes or objects, described by text.
Following Neculai~\etal~\cite{neculai2022probabilistic}, who use COCO~\cite{lin2014microsoft} for composed image retrieval, COCO validation set (5,000 images) is used for evaluation. To create a query for each image, we randomly crop one object and mask its background using its instance mask. The list of object classes, including the class of the query image, in the image is used as text specification. Given the reference image and class list, we compose query by following (b) in Sec~\ref{sec:compose}, e.g., \texttt{a photo of [*], [car], [cat], and dog}.

\noindent\textbf{(c). Scene manipulation by text description.}
CIRR~\cite{liu2021image} (Fig.~\ref{fig:ex_cirr}) is employed to evaluate image manipulation described by text. The modification from the reference to the target image is given as a sentence. We follow the standard evaluation protocol and compose query text following (c) in Sec~\ref{sec:compose}. We report the results on the test split only in Table~\ref{tab:cirr_test}, the validation split is used in other analysis.

\begin{figure}[t]
    \centering
    \includegraphics[width=0.44\textwidth]{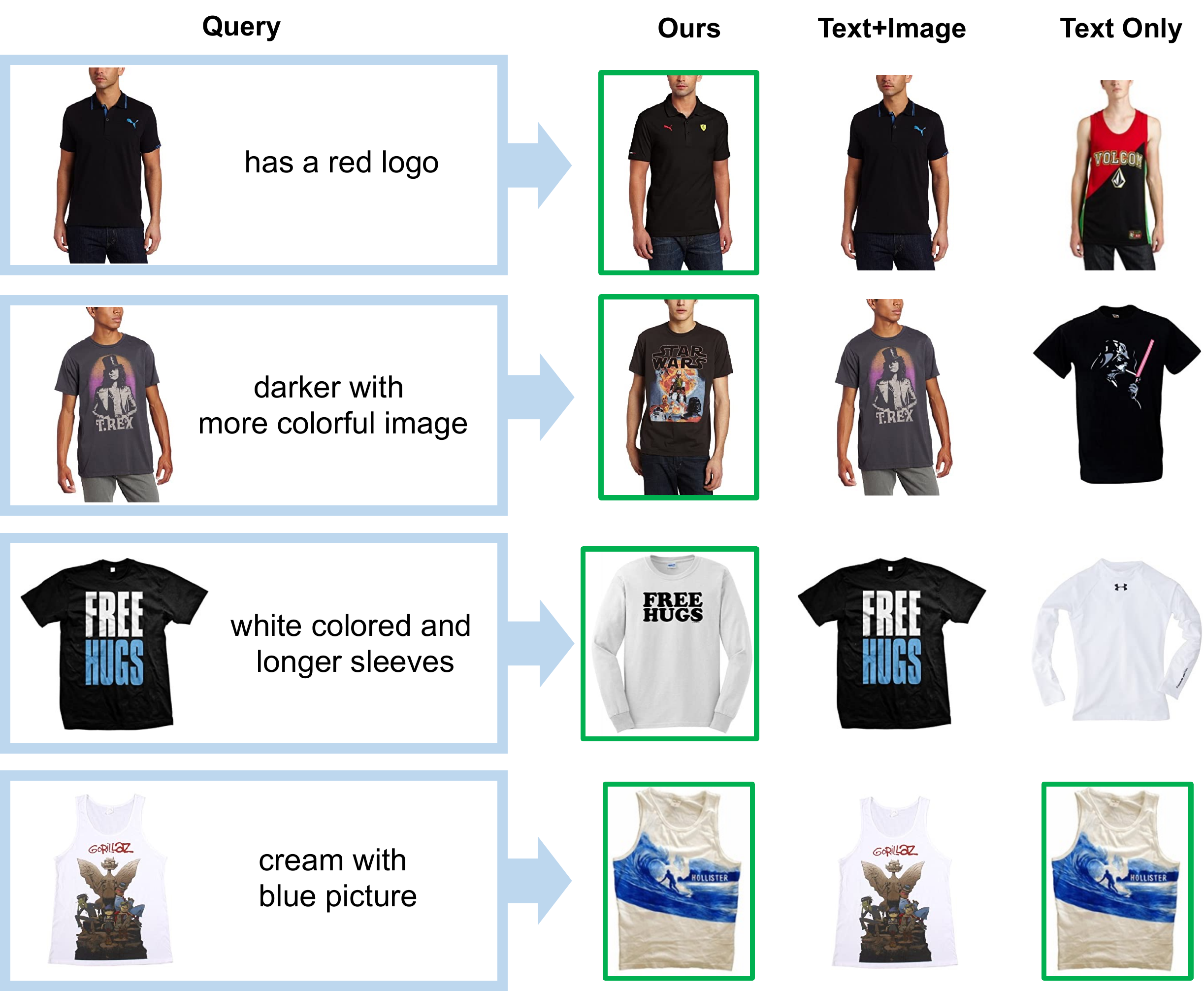}
    \vspace{-3mm}
    \caption{\textbf{Retrieval examples in Fashion-IQ.} The search is done in the validation split of Fashion-IQ dataset. Note that Text + Image baseline returns the reference image. The baseline is likely to put more importance on image features although the text and image features are averaged with an equal weight.}
    \label{fig:ex_fashion}
\end{figure}

\noindent\textbf{(d). Fashion attribute manipulation.}
Fashion-IQ~\cite{wu2021fashion} is employed to evaluate the manipulation of fashion images. The attribute manipulation is given as a sentence.
Similarly to CIRR, we follow the standard evaluation protocol and compose query text following (c) in Sec~\ref{sec:compose}. The validation set is used for evaluation following previous work~\cite{baldrati2022effective}.

\textbf{Zero-shot baselines.}
We provide three zero-shot baselines to fairly compare with our approach. 
\begin{itemize}
    \item \textbf{Image Only.} This baseline retrieves images by computing similarity between target image features and  query image features. 
    \vspace{-2mm}
    \item \textbf{Text Only.} This baseline employs text features only to compute similarity with target images. 
    \vspace{-2mm}
    \item \textbf{Image and Text Baseline.} This baseline utilizes the mean of image and text features as query features. Before taking the average, both image and text features are normalized to have a unit $L_2$-norm.
     \vspace{-2mm}
\end{itemize}
\begin{table*}[t]
  \centering \scalebox{1.0}{
  \begin{tabular}{lcrr|rr|rr|rr} 
  \toprule
  \multicolumn{2}{c}{}     &   \multicolumn{2}{c}{Dress} & \multicolumn{2}{c}{Shirt}& \multicolumn{2}{c}{TopTee} & \multicolumn{2}{c}{Average}\\
   \cmidrule(lr){3-4}
  \cmidrule(lr){5-6}
  \cmidrule(lr){7-8}
  \cmidrule(lr){9-10}
  \multicolumn{1}{l}{Supervision} & \multicolumn{1}{c}{Methods} & R10 & R50& R10 & R50& R10 & R50& R10 & R50\\  \midrule
  \multirow{4}{*}{{\textsc{Zero-shot}}}
  & Image-only  &5.4&13.9&9.9&20.8&8.3&17.7&7.9&17.5\\
  &  Text-only  &13.6&29.7&18.9&31.8&19.3&37.0&17.3&32.9\\
  &  Image$+$Text& 16.3&33.6&21.0&34.5&22.2&39.0&19.8&35.7\\
  &  \ours   &  \textbf{20.0}&\textbf{40.2}&\textbf{26.2}&\textbf{43.6}&\textbf{27.9}&\textbf{47.4}&\textbf{24.7}&\textbf{43.7}\\

\midrule
\multicolumn{1}{c}{CIRR}  &Combiner~\cite{baldrati2022effective} &17.2&37.9&23.7&39.4&24.1	&43.9&21.7&40.4 \\
\multicolumn{1}{c}{Fashion-IQ}&Combiner~\cite{baldrati2022effective}&30.3&54.5&37.2&55.8&39.2&61.3&35.6&57.2\\\midrule
\multicolumn{1}{c}{Fashion-IQ}&Combiner$^{*}$~\cite{baldrati2022effective}&31.6&56.7&36.4&58.0&38.2&62.4&35.4&59.0\\

\multicolumn{1}{c}{Fashion-IQ}& CIRPLANT~\cite{liu2021image}&17.5&40.4&17.5&38.8&21.6&45.4& 18.9 &41.5\\
\multicolumn{1}{c}{Fashion-IQ}&ALTEMIS~\cite{delmas2022artemis}&27.2&52.4&21.8&43.6&29.2&54.8&26.1&50.3\\
\multicolumn{1}{c}{Fashion-IQ}&MAAF~\cite{dodds2020modality}&23.8&48.6&21.3&44.2&27.9&53.6&24.3&48.8\\
\bottomrule
  \end{tabular}}
   \vspace{-3mm}
  \caption{\textbf{Results on Fashion-IQ validation set.} Combiner$^{*}$ is the result reported by the authors using ResNet50x4 as a backbone.}
  \label{tab:fashion}
\end{table*}
\begin{figure*}[!htb]
     \centering
     \begin{subfigure}[b]{0.24\textwidth}
         \centering
         \includegraphics[width=\textwidth]{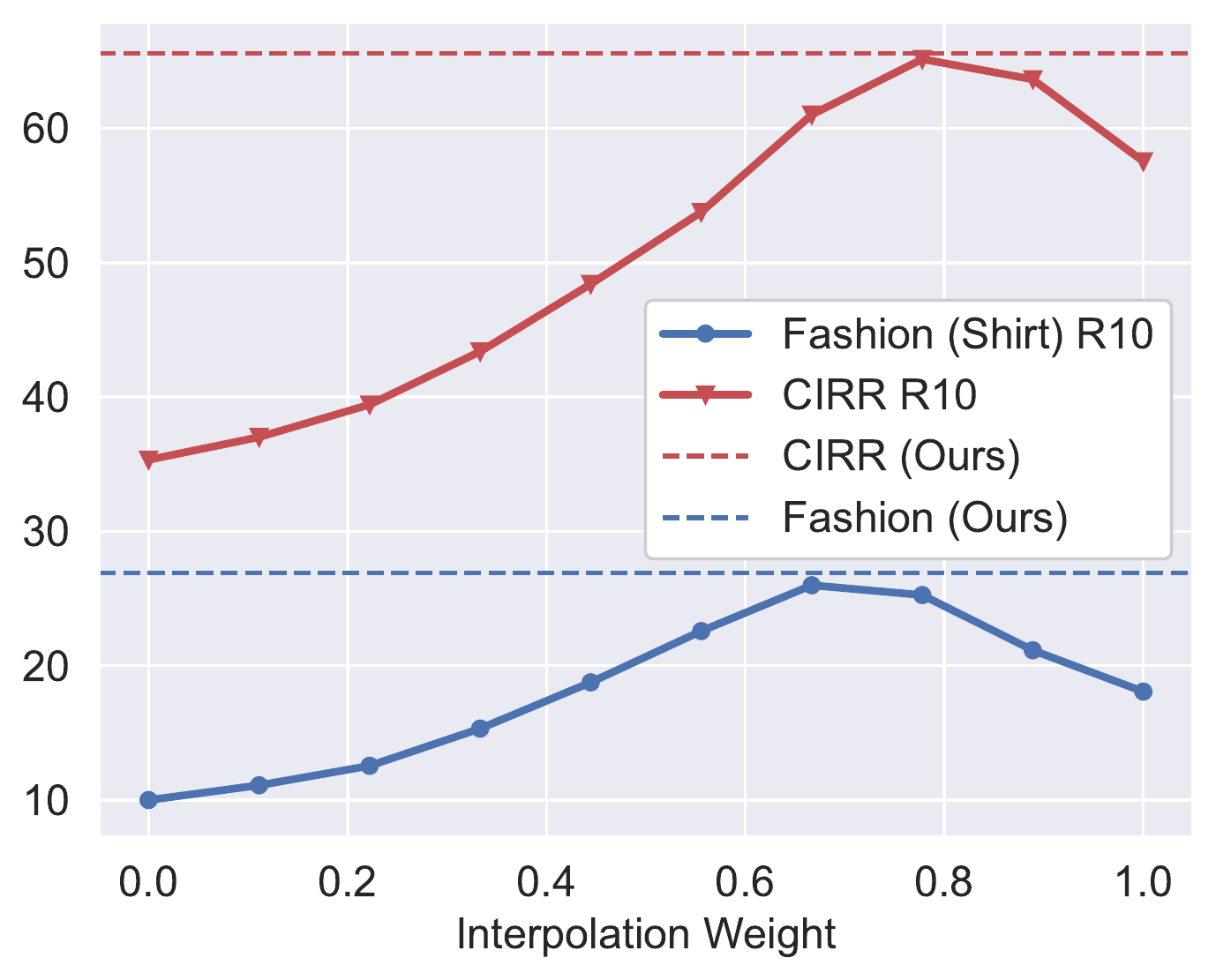}
         \caption{}
         \label{fig:weight_study}
     \end{subfigure}
     \begin{subfigure}[b]{0.24\textwidth}
         \centering
         \includegraphics[width=\textwidth]{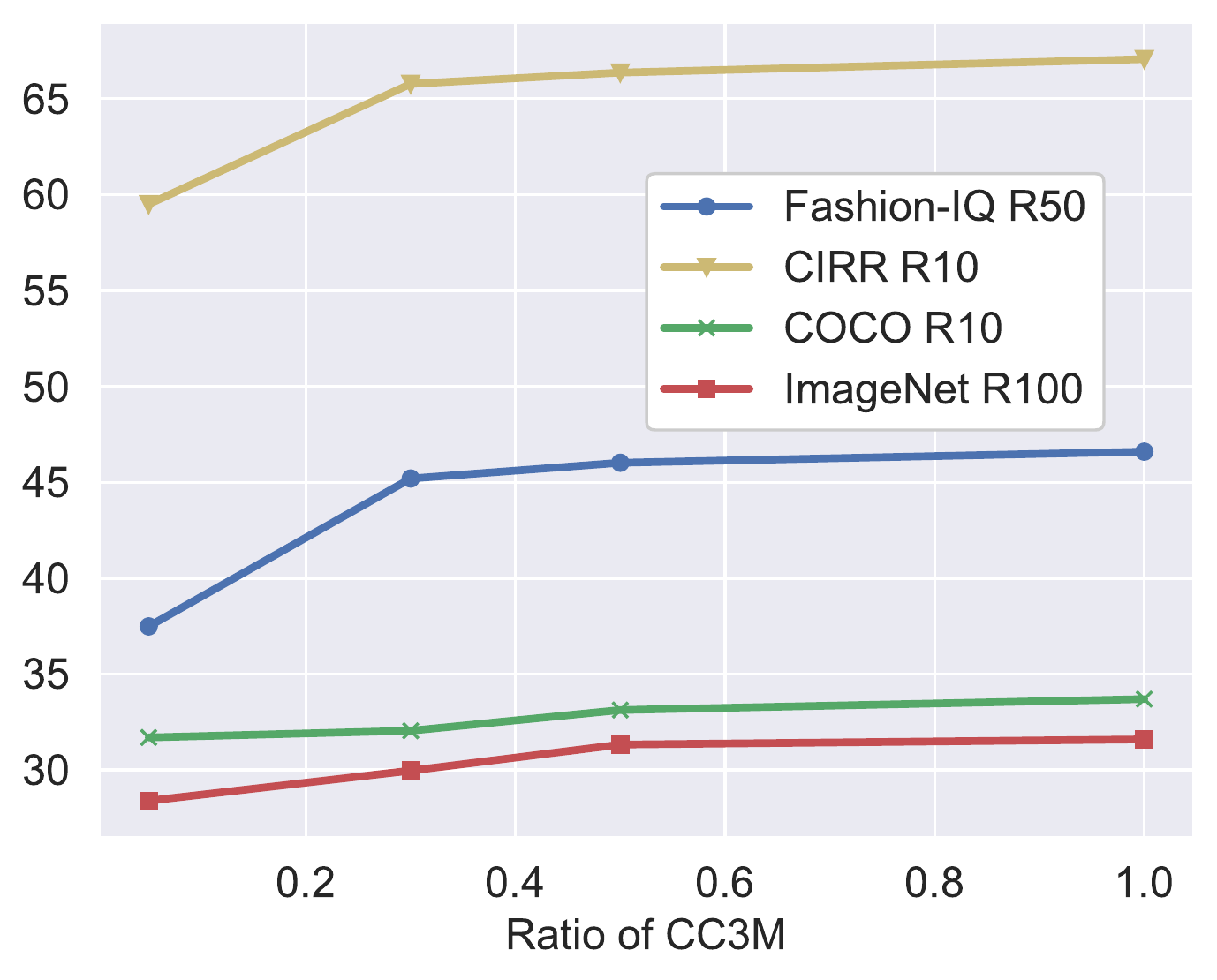}
         \caption{}
         \label{fig:cc3m_num_sample}
     \end{subfigure}
     \hfill
     \begin{subfigure}[b]{0.24\textwidth}
         \centering
         \includegraphics[width=\textwidth]{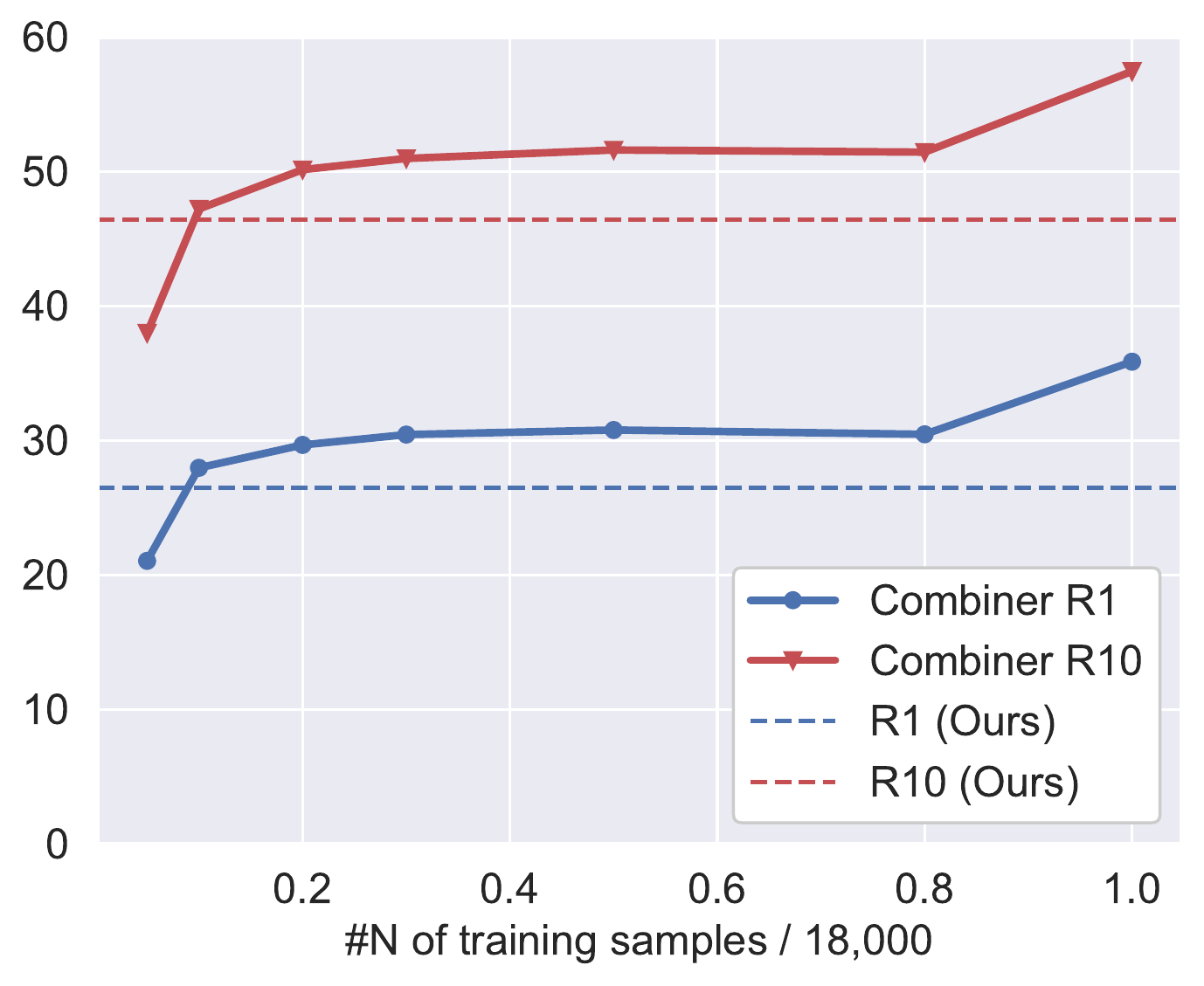}
         \caption{}
         \label{fig:fashion_num_sample}
     \end{subfigure}
     \hfill
     \begin{subfigure}[b]{0.24\textwidth}
         \centering
         \includegraphics[width=\textwidth]{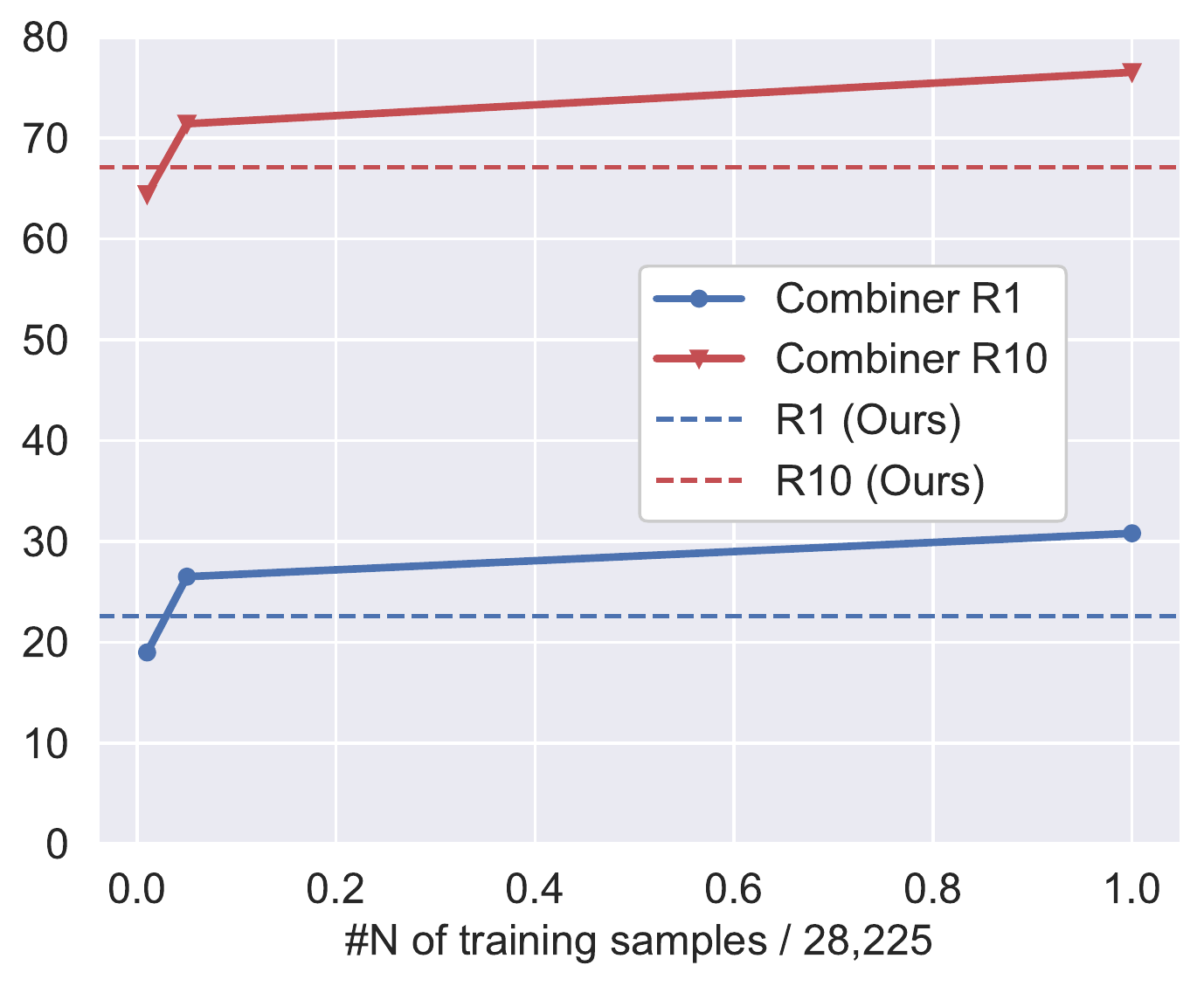}
         \caption{}
         \label{fig:cirr_num_samples}
     \end{subfigure}
      \vspace{-3mm}
        \caption{(a): Analysis on the importance weight to average text and image features. (b): Performance change by the number of CC3M samples used to train mapping network. (c): Performance change of the supervised baseline (Combiner) by the number of supervised training samples in Fashion-IQ dataset. (d): Performance change of the supervised baseline (Combiner) by the number of supervised training samples in CIRR dataset.}
        \label{fig:three_graphs}
\end{figure*}

\textbf{Supervised baselines.} To better understand the performance of zero-shot methods, we compare with baselines trained with labeled triplets of a CIR dataset. We implement Combiner~\cite{baldrati2022effective} model with a CLIP ViT-L/14 backbone following the author's code.\footnote{\url{https://github.com/ABaldrati/CLIP4CirDemo}} We train the model with the training split of CIRR or Fashion-IQ; approximately, CIRR has 28,000 training triplets and Fashion-IQ includes 18,000. We also show the reported performance of several CIR methods for CIRR and Fashion-IQ.

\subsection{Main Results}
Tables~\ref{tab:imgnet}-\ref{tab:fashion} present quantitative results, Fig.~\ref{fig:ex_imgnet}-\ref{fig:ex_fashion} illustrate qualitative results. In the experiment on domain conversion using ImageNet (Table~\ref{tab:imgnet}), \ours outperforms all of the baselines with a large margin. This indicates the superiority of \ours in composing domain words and image features. Fig.~\ref{fig:ex_imgnet} demonstrates that \ours captures both the domain and image representations well. 

In the experiment on object composition using COCO (Table~\ref{tab:coco}), \ours outperforms the zero-shot baselines.  Combiner~\cite{baldrati2022effective} trained on Fashion-IQ performs better than \ours while the model trained on CIRR performs worse than \ours. A model trained on one dataset is not necessarily transferable to other datasets because each composed retrieval dataset can require different dataset bias, e.g., relative importance of image features and caption features. 

On scene and fashion attribute manipulation datasets, e.g., CIRR (Table~\ref{tab:cirr_test}) and Fashion-IQ (Table~\ref{tab:fashion}), \ours outperforms zero-shot baselines and some of the supervised approaches. 
For these datasets, we observe that the captions alone can be informative enough to retrieve the correct target images (e.g., last row in Fig.~\ref{fig:ex_cirr}), and that some reference images are not very relevant to the target image. 
Then, the supervised model can learn the dataset-specific bias on how to combine the two modalities. For example, the model can prioritize learning from text or ignoring the reference image if the image and text are not related well.
We hypothesize that this is affecting the transferability of the supervised models. 
More detailed analysis is provided in the following section.

\subsection{Analysis}
\label{sec:analysis}
\textbf{Does the pseudo language token capture the image information?}
To analyze the information in the estimated token, we conduct evaluation using CC3M validation 13,164 images. Concretely, we evaluate whether the language embedding generated by the pseudo token can retrieve the input image. Recall at top 1 and 5 is 99.8 and 100.0 respectively, indicating that the token captures the unique image features very well.

\begin{figure*}
    \centering
    \includegraphics[width=\textwidth]{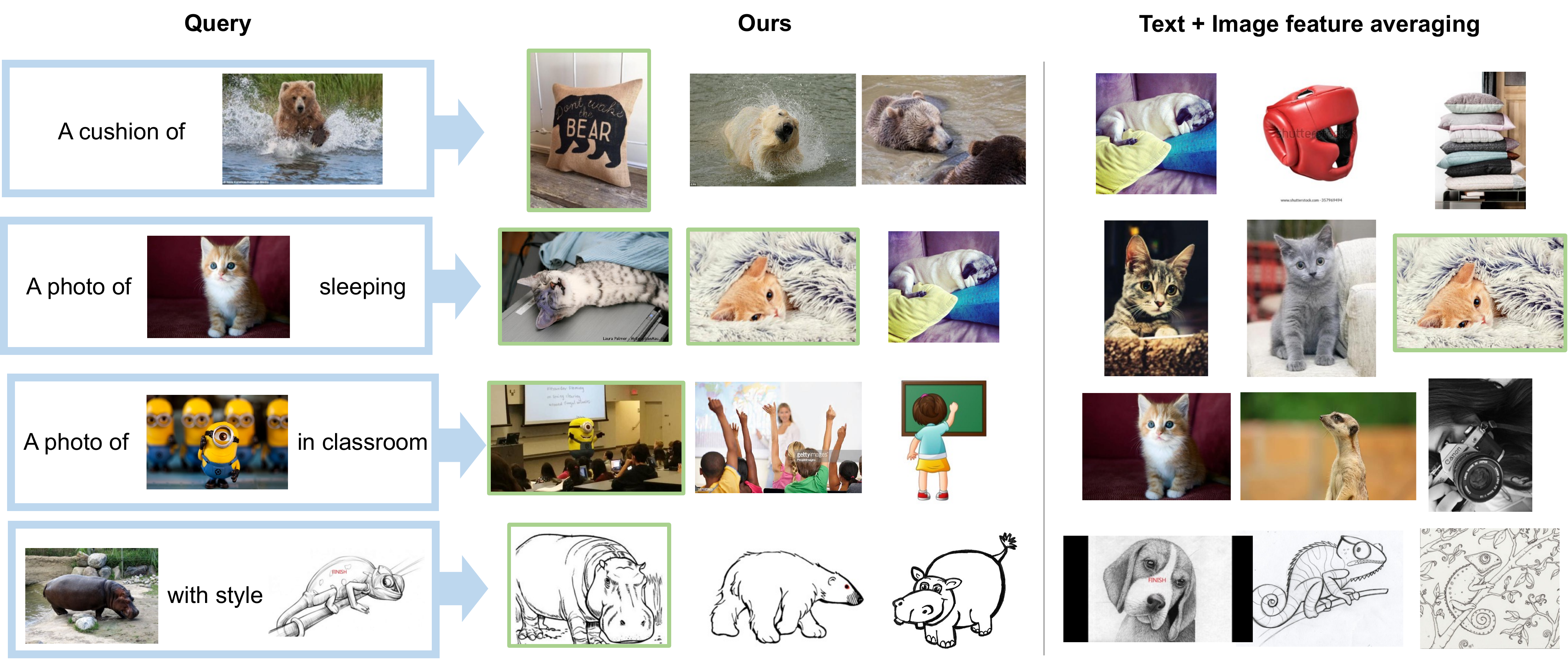}
     \vspace{-5mm}
    \caption{Top-3 examples retrieved from CC3M validation set (13,000 images). Correct retrievals are highlighted with green outline.}
    \label{fig:interesting}
\end{figure*}
\begin{figure}
    \centering
    \includegraphics[width=0.48\textwidth]{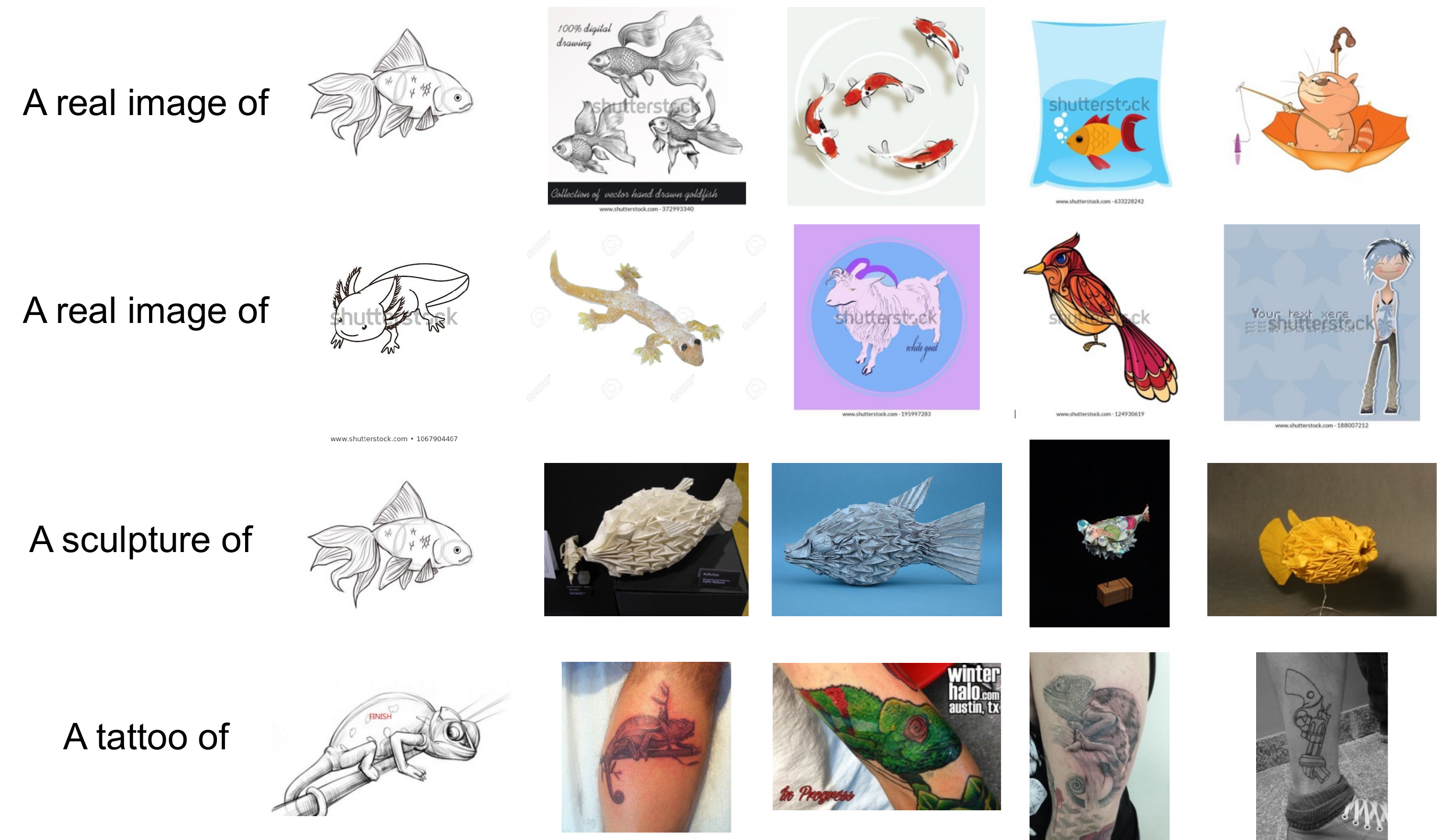}
     \vspace{-4mm}
    \caption{Retrieval with sketch-style images. We qualitatively find that it can be hard to handle sketch-to-real image retrieval (Top two). }
    \label{fig:failures}
\end{figure}
\textbf{Insights on existing composed retrieval datasets.}
The results on CIRR and Fashion-IQ (Table~\ref{tab:cirr_test} and ~\ref{tab:fashion}) imply that some target images can be correctly retrieved just by using text. We find that reference images can be irrelevant to text query or text query is informative enough to search target image. This relative importance of reference image and text is a dataset-specific bias. We investigate the importance in Fig.~\ref{fig:weight_study}, where we produce a query embedding by varying the interpolation weight, $w$, in computing the average of text $t$ and image features $v$, i.e., $q = w * t + (1-w) * v$. The result indicates that the optimal weight can be unique to each dataset and the performance is sensitive to the weight parameter. A supervised method can learn the relative importance from labeled triplets, but it is hard to learn for a zero-shot method. 

\textbf{Comparison to supervised baseline with fewer training samples.}
Fig.~\ref{fig:fashion_num_sample} and \ref{fig:cirr_num_samples} show the performance of the supervised baseline (Combiner~\cite{baldrati2022effective}) with fewer training samples. Overall, our approach outperforms the baseline if fewer than 1,000 triplets are given as training samples.

\textbf{Number of training samples to train the mapping network.}
In Fig. ~\ref{fig:cc3m_num_sample}, we investigate the performance w.r.t the number of data (from CC3M) used to train the mapping network. We observe that using only 10\% degrades the performance while the use 50\% data is comparable to all data.

\textbf{More qualitative examples.}
Fig.~\ref{fig:interesting} shows retrieval results using CC3M validation data. 
Our model can compose images with adjective (second) or a place (third) and read the style given as an image (bottom). 

\textbf{Failure cases.}
One of the popular applications is sketch-based image retrieval~\cite{liu2017deep}, where the user draws a sketch to retrieve natural images, as shown in Fig.~\ref{fig:failures}.
Qualitatively, retrieving an image from a non-natural image domain is likely to be successful (bottom two) while retrieving natural images from sketch images is not trivial (top two). 
Prompts other than \texttt{a real image of} do not work well for this task either. Natural images are not described with a domain word in pre-training image-caption dataset while other domains, e.g., art, cartoon, sketch and origami, are often described with corresponding domain name. An appropriate domain description for natural images may not exist. 

\section{Conclusion}
\vspace{-2mm}
In this paper, we present a novel task, zero-shot composed image retrieval, and the first method to approach the problem. We propose to employ a pre-trained CLIP model that treats an image as a text token so that the language encoder can flexibly compose the image features and text description. 
We perform a comprehensive analysis on four datasets, demonstrating that our approach, \ours, shows strong generalization across diverse CIR tasks, with performance being on-par with or better than several recent CIR methods that require labeled training data. 

\noindent \textbf{Acknowledgment.}
We thank Zizhao Zhang for their helpful feedback on the manuscript. This work was supported in part by DARPA LwLL.

{\small
\bibliographystyle{ieee_fullname}
\bibliography{egbib}
}

\clearpage
\newpage
\setcounter{section}{0}
\setcounter{figure}{0}
\setcounter{table}{0}
\def\thesection{\Alph{section}}
\def\thetable{\Alph{table}} 
\def\thefigure{\Alph{figure}}

\section{Additional Discussion}
\textbf{Discussion on DreamBooth~\cite{ruiz2022dreambooth} and Textual Inversion~\cite{gal2022image}}
From the application perspective, these works generate images with the user's intent while our approach retrieves images given the intent. Besides the difference from the application perspective, there are few critical differences. First, our approach needs only unlabeled images, while their approaches rely on a set of images containing the same object. Second, our method works in real-time as it requires single forward pass of visual encoder, mapping network and text encoder at inference time. On the other hand, both Textual Inversion and DreamBooth are far from being a real-time as they require more than thousands of gradient steps to invert or finetune per set of images.

\section{Experimental Details}
\textbf{Evaluation Dataset.}
Table~\ref{tab:dataset_details} describes the details of the dataset, i.e., number of query images and candidate images used for evaluation. The evaluation datasets are preprocessed as explained in the main paper.
\begin{table}[h]
  \centering 
  \scalebox{0.9}{
  \begin{tabular}{c|c|c} 
  \toprule
  \cmidrule{1-3}
  Dataset & Query images & Candidate images\\ 
  \midrule
  ImageNet &10,000& 16,983\\
  COCO &4,766&4,766\\ 
  CIRR (test) &4,148&2,315\\
  Fashion (Dress) & 2,017 & 3,817\\
  Fashion (Shirt) &2,038&6,346\\
   Fashion (TopTee) & 1,961&5,373\\
  \bottomrule
  \end{tabular}}
   \vspace{-3mm}
   \caption{The number of images used for evaluation in each dataset.}
  \label{tab:dataset_details}
\end{table}

\textbf{Mapping network design.}
Table~\ref{tab:network_design} summarizes the mapping network architecture we employ. In the next section, we give the study on the choice of the architecture. 
\begin{table}[h]
 \centering 
  \scalebox{0.9}{
\begin{tabular}{c|cccc}
 \toprule
Layer & Module\\
\hline
Output & nn.Linear(512, 768)\\\hline
ReLU2&nn.ReLU\\\hline
Dropout2&nn.Dropout(0.1)\\\hline
FC2&nn.Linear(512, 512)\\\hline
ReLU1&nn.ReLU\\\hline
Dropout1&nn.Dropout(0.1)\\\hline
FC1&nn.Linear(512, 512)\\
\bottomrule
\end{tabular}
}
\vspace{-3mm}
\caption{Pytorch-style\cite{paszke2019pytorch} model description of the mapping network. The output is fed into the langauge encoder.}

\label{tab:network_design}
\end{table}

\textbf{Images used for qualitative examples.}
In qualitative examples, we exclude images that can recognize the identity of a person. 
In the third query in Fig. 8, we employ an image\footnote{\url{https://www.flickr.com/photos/enerva/9068467267} licensed with CC-BY 2.0.}, which is not included in CC3M validation set.

\begin{table}
 \centering 
  \scalebox{0.8}{
\begin{tabular}{c|cccc}
 \toprule
\multirow{2}{*}{\makecell{Model \\ Description}}& \multirow{2}{*}{\makecell{ImageNet \\ R50}}    & \multirow{2}{*}{\makecell{COCO \\ R10}} & \multirow{2}{*}{\makecell{CIRR \\ R10}} & \multirow{2}{*}{\makecell{Fashion \\ R50}} \\
                  &&&\\\hline
Default ($L = 3$, $h_{d}=512$)&\textbf{23.2}&33.4&65.4&43.7\\
$L = 3$, $h_{d}=4096$&22.4&\textbf{33.9}&\textbf{67.6}&\textbf{45.3}\\
$L = 5$, $h_{d}=512$ &22.3&32.3&65.1&42.6\\
 Linear only ($L = 2$, $h_{d}=512$)&22.1&33.4&58.8&42.4\\ \hline
Best zero-shot baseline& 11.2 & 26.6 & 56.7 & 35.7    \\
\bottomrule
\end{tabular}
}
\vspace{-3mm}
\caption{Analysis of the design of the mapping network. The top row is the model used in the main paper. The bottom is the score of the zero-shot baseline, which performs the best of three zero-shot baselines in each dataset.}
\label{tab:architecture}
\end{table}

\begin{table*}[t]
  \centering 
  \scalebox{0.8}{
  \begin{tabular}{c|cccccccccc} 
  \toprule
  \cmidrule{1-5}
  \multirow{2}{*}{Training Dataset} &  \multicolumn{2}{c}{ImageNet} & \multicolumn{2}{c}{COCO} &
  \multicolumn{2}{c}{CIRR} &\multicolumn{2}{c}{Fashion}      \\ 
  \cmidrule(lr){2-3}
  \cmidrule(lr){4-5}
  \cmidrule(lr){6-7}
  \cmidrule(lr){8-9}
  &R10&R50&R1&R10&R1&R10&R10&R50\\
  \midrule

  CC3M    &10.1$\pm$1.5&23.2$\pm$1.1&11.5$\pm$0.2&33.4$\pm$0.3&22.2$\pm$0.6&65.4$\pm$1.3&24.7$\pm$2.1&43.7$\pm$3.4\\
  CC12M &8.9$\pm$1.4&21.4$\pm$0.6&12.1$\pm$1.1&33.9$\pm$1.0&22.1$\pm$2.5&64.5$\pm$3.4&25.1$\pm$0.7&43.4$\pm$1.2\\
  \bottomrule
  \end{tabular}}
  \vspace{-3mm}
  \caption{\textbf{Study on the dataset to train mapping network.} We report the results averaged over three runs and its standard deviation. Note that we report the results on validation set for CIRR.}
  \label{tab:cc3m_vs_cc12m}
\end{table*}

\begin{table}
  \centering \scalebox{1.0}{
  \begin{tabular}{llrrrrr} 
  \toprule
 
\multirow{2}{*}{Model}  & \multirow{2}{*}{Methods}  &   \multicolumn{2}{c}{COCO} & \multicolumn{2}{c}{Fashion}\\
   \cmidrule(lr){3-4}
  \cmidrule(lr){5-6}
 &  & R5 & R10&R10 & R50\\  \midrule
\multirow{2}{*}{Fine-tuned}  &  Text-only &14.3&22.0&19.4&37.1              \\
  &  Image$+$Text & 22.5&29.0&20.9&36.9               \\
  \cmidrule{1-6}
\multirow{3}{*}{Original} & Text Only&	15.7&23.5&17.3&32.8\\
&Image + Text&20.2&26.6&19.8&35.7\\
  &   Ours &24.8&33.4&24.7&43.7\\
  \bottomrule
  \end{tabular}}
  \caption{Baseline results of CLIP fine-tuned on CC3M (top two).}
  \label{tab:cc3m_tuned}
\end{table}
\begin{table}[t]
  \centering 
  \scalebox{0.9}{
  \begin{tabular}{c|c|cccc} 
  \toprule
  \cmidrule{1-5}
  \multicolumn{1}{l}{Model} & \multicolumn{1}{c}{Methods} & R1& R5&R10&R50\\ 
  \midrule
   \multirow{3}{*}{{\textsc{BLIP}}} & Image-only    &7.2&25.6&36.6&62.4
\\ 
   & Text-only   &\textbf{25.1}&52.0&62.4&82.7\\
   & Image$+$Text  &16.5&47.2&61.3&86.8
\\
\cmidrule{1-6}
   \multirow{4}{*}{{\textsc{CLIP}}} & Image-only    &7.5&25.1&	35.5&59.8\\ 
   & Text-only   &20.8&46.2&57.0&78.8\\
   & Image$+$Text  &13.2&36.6&50.5&78.1\\
   & \ours &22.6&\textbf{52.6}&\textbf{66.6}&\textbf{87.3}\\
  \bottomrule
  \end{tabular}}
  \caption{Comparison between CLIP and BLIP ViT-L/14 model~\cite{li2022blip_reb}. Evaluation on CIRR validation set.}
  \label{tab:blip_cirr}
\end{table}

\section{Additional Experiments}
\textbf{Analysis on the architecture design.}
Table~\ref{tab:architecture} shows the study on the design of the mapping network. There are two observations: (1) no variants outperform the default model across all datasets, (2) removing non-linear activation can significantly reduce the performance (Linear only model). In order to faithfully predict the pseudo language token, the mapping network needs to be expressive to a certain degree. We also try other variants, e.g., varying dropout rate, but we do not see clear improvements.

\textbf{Training dataset for the mapping network.}
Table~\ref{tab:cc3m_vs_cc12m} describes the comparison between the model trained on CC3M and CC12M, where CC12M is 4 times larger than CC3M approximately. We do not see the clear advantage of using CC12M, which indicates that CC3M is large enough to train the mapping network.

 \textbf{Comparison with CLIP fine-tuned on Conceptual Caption (CC3M).} 
In Table~\ref{tab:cc3m_tuned}, we show that fine-tuning CLIP on CC3M improves the baseline performance. However, these baselines still fall short to our approach.

 \textbf{Comparison between CLIP and BLIP.} 
In Table~\ref{tab:blip_cirr}, we show the comparison between CLIP and BLIP~\cite{li2022blip_reb} ViT-L/14 model. Text-only result of BLIP outperform that of CLIP, showing that difference in pre-training can result in the significant difference in the performance.

\begin{figure*}[t]
    \centering
     \vspace{-3mm}
    \includegraphics[width=\textwidth]{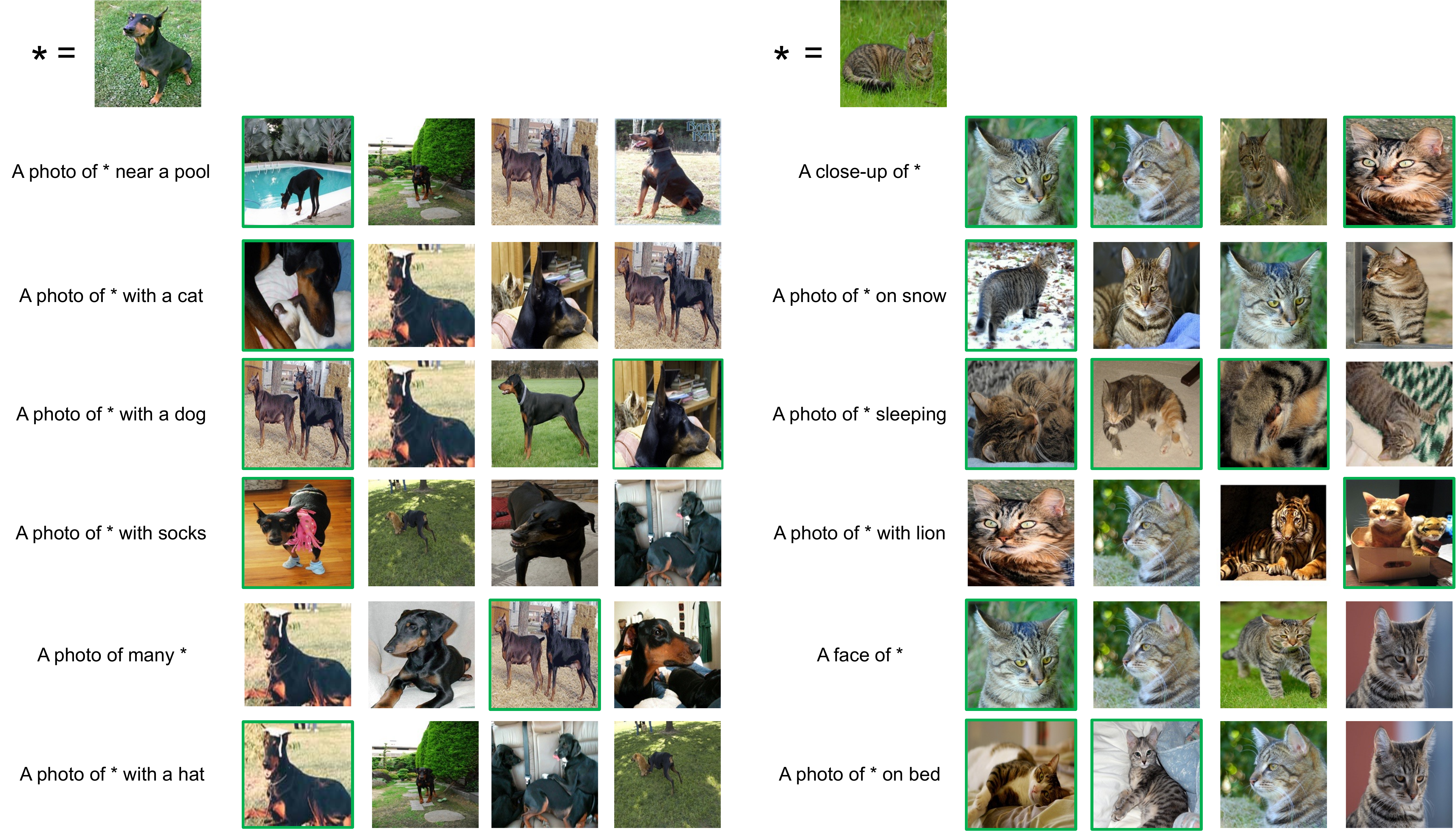}
    \vspace{-7mm}
    \caption{Image retrieval examples in the synset of n02107142 (doberman) and n02123159 (tiger cat). Note that the search is performed on 50 images from each category. Target images are highlighted with green outline.} 
    \label{fig:ex_fine}
\end{figure*}

\begin{figure*}[t]
    \centering
     \vspace{-3mm}
    \includegraphics[width=\textwidth]{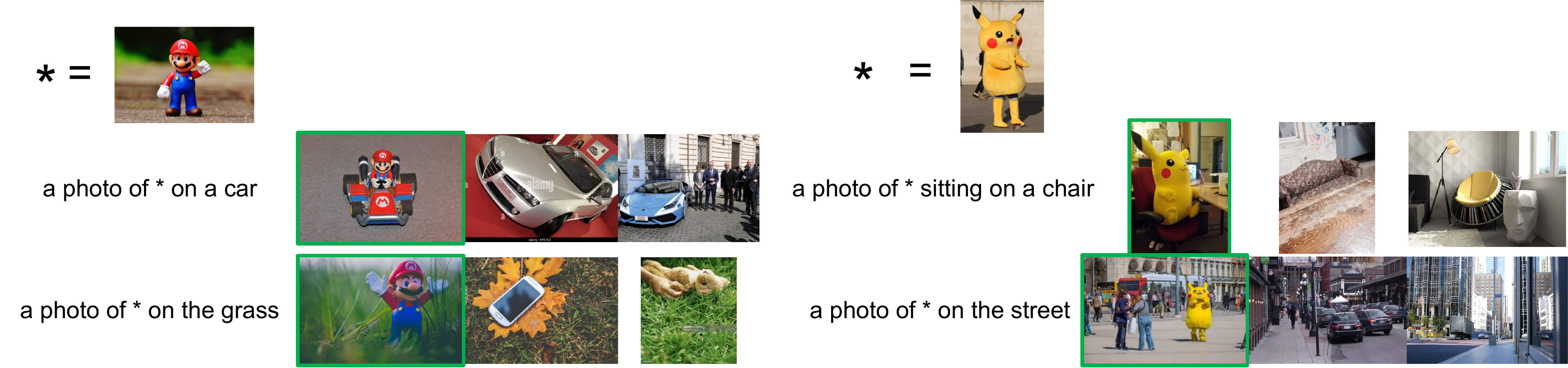}
     \vspace{-7mm}
    \caption{Qualitative retrieval results. Query images are from \href{https://images.rawpixel.com/image_1300/czNmcy1wcml2YXRlL3Jhd3BpeGVsX2ltYWdlcy93ZWJzaXRlX2NvbnRlbnQvbHIvZnJtYXJpb19maWdfcGxheV9uaW50ZW5kby1pbWFnZS1rejJkem1yNC5qcGc.jpg}{rawpixel} and \href{https://commons.wikimedia.org/wiki/File:Pokemon_(31599837283).jpg}{wikimedia}. Top-1 images are downloaded from
    \href{https://www.flickr.com/photos/30478819@N08/19701406723}{flickr},  \href{https://images.rawpixel.com/image_800/cHJpdmF0ZS9zdGF0aWMvaW1hZ2Uvd2Vic2l0ZS8yMDIyLTA0L2xyL2ZybmF0dXJlX3RveV9jaGlsZHJlbl8xMjgzNTY4LWltYWdlLWt6MmU3cXVsLmpwZw.jpg?s=_96SBsW2K7UveZwQ3NJnybKApRptjZnjKwO5cpaJ3VA}{rawpixel},
    \href{https://www.flickr.com/photos/photographingtravis/16025479579}{flickr}, and 
    \href{https://commons.wikimedia.org/wiki/File:Pikachu_(34296520333).jpg}{wikimedia} (top to bottom and left to right). Note that these images are licensed by either CC-BY 2.0, CC BY-SA 2.0, or CC0 1.0 Universal. We create a candidate set by these images and CC3M validation split. Target images are highlighted with green outline.}
    \label{fig:ex_pikachu}
\end{figure*}

\textbf{Additional qualitative examples.}
Fig.~\ref{fig:ex_fine} shows retrieval examples of ImageNet. Both the query and candidate images are from the same synset of one category. We can see that target images are within top-4 in these examples. 
Fig.~\ref{fig:ex_pikachu} shows retrieval results tested on the images downloaded from the web. In this evaluation, all of the top-1 images are included in the candidate set, which indicates that the composed representations can express both the characteristics of an object, specified by an image, and the attribute from language.

\end{document}